\documentclass[letterpaper]{article} 
\usepackage{aaai24}  
\usepackage{times}  
\usepackage{helvet}  
\usepackage{courier}  
\usepackage[hyphens]{url}  
\usepackage{graphicx} 
\usepackage{natbib}  
\usepackage{caption} 
\usepackage{algorithm}
\usepackage{algorithmic}

\usepackage{newfloat}
\usepackage{listings}
\usepackage{multirow}
\usepackage{color}
\DeclareCaptionStyle{ruled}{labelfont=normalfont,labelsep=colon,strut=off} 
\floatstyle{ruled}
\newfloat{listing}{tb}{lst}{}
\floatname{listing}{Listing}
\title{Shadow Generation with Decomposed Mask Prediction and Attentive Shadow Filling}
\author {
    Xinhao Tao\textsuperscript{\rm 1},
    Junyan Cao\textsuperscript{\rm 1},
    Yan Hong\textsuperscript{\rm 2},
    Li Niu\textsuperscript{\rm 1}\thanks{Corresponding author.}
}
\affiliations {
    \textsuperscript{\rm 1}Shanghai Jiao Tong University \\
    \textsuperscript{\rm 2}Tiansuan Lab, Ant Group \\
    \{taoxinhao, Joy\_C1, ustcnewly\}@sjtu.edu.cn, yanhong.sjtu@gmail.com
}

\begin{document}

\maketitle

\begin{abstract}
Image composition refers to inserting a foreground object into a background image to obtain a composite image. In this work, we focus on generating plausible shadows for the inserted foreground object to make the composite image more realistic. To supplement the existing small-scale dataset, we create a large-scale dataset called RdSOBA with rendering techniques. Moreover, we design a two-stage network named DMASNet with \textbf{d}ecomposed \textbf{m}ask prediction and \textbf{a}ttentive \textbf{s}hadow filling. Specifically, in the first stage, we decompose shadow mask prediction into box prediction and shape prediction. In the second stage, we attend to reference background shadow pixels to fill the foreground shadow. Abundant experiments prove that our DMASNet achieves better visual effects and generalizes well to real composite images.

\end{abstract}

\section{Introduction} \label{sec:intro}
Image composition refers to cutting out a foreground object and pasting it on another background image to acquire a composite image, which could benefit plenty of applications in art, movie, and daily photography \cite{ic_useful1,ic_useful2,ic_useful3}. However, the quality of composite images could be significantly compromised by the inconsistency between foreground and background, including appearance, geometric, and semantic inconsistencies. 
In recent years, many deep learning models \cite{ic_full1,ic_full2,ic_full3} have endeavored to tackle different types of inconsistencies in composite images, but only a few works \cite{sgrnet} focused on the shadow inconsistency, which is a crucial aspect of appearance inconsistency. In this work, we aim to cope with the shadow inconsistency, \emph{i.e.}, generating plausible shadow for the foreground object to make the composite image more realistic.

\begin{figure}[t]
\centering
\includegraphics[width=1.0\linewidth]{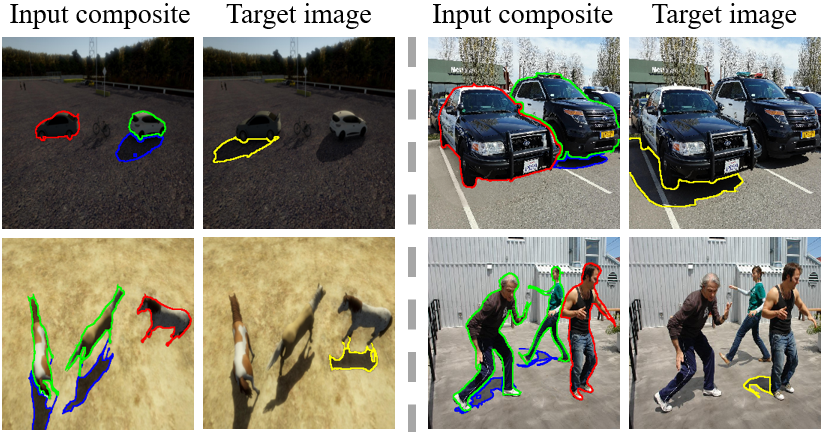}
\caption{The example data for shadow generation task. The left two examples are from our RdSOBA dataset and the right two examples are from DESOBA dataset \cite{sgrnet}. In each example, the background object (\emph{resp.}, shadow) mask is outlined in {\color{green}green} (\emph{resp.}, {\color{blue}blue}) and the foreground object (\emph{resp.}, shadow) mask is outlined in {\color{red}red} (\emph{resp.}, {\color{yellow}yellow}). }
\label{fig:example}
\end{figure}

For the shadow generation task, the used data are exhibited in Figure \ref{fig:example}. Given a composite image $I_c$ without foreground shadow, the foreground object mask $M_{fo}$, and the masks of background object-shadow pairs $\{M_{bo},M_{bs}\}$, our goal is generating $\hat{I}_g$ with foreground shadow. To solve this problem, SGRNet~\cite{sgrnet} released the first synthetic dataset DESOBA for real-world scenes. The data in DESOBA can be summarized as tuples of the form $\{I_c,M_{fo},M_{fs},M_{bo},M_{bs},I_g\}$. The way to obtain pairs in DESOBA is as follows. First, they manually remove the shadows in real images to get shadow-free images. Then, they replace one foreground shadow region in a real image (ground-truth image) with the counterpart in its shadow-free version, yielding a synthetic composite image without foreground shadow.  Due to the high cost of manual shadow removal, the scale of DESOBA dataset is very limited (\emph{e.g.}, 1012 ground-truth images and 3623 tuples). Nevertheless, deep learning models require abundant training data. 

To supplement DESOBA dataset, we create a large-scale dataset RdSOBA using rendering techniques. We first collect lots of 3D objects and 3D scenes, then place a group of 3D objects in the 3D scene. By using rendering techniques, we can get the images without object shadows and the images with object shadows without manual effort. Finally, we obtain 114,350 ground-truth images and 280,000 tuples. The details of dataset construction can be found in Section \ref{sec:RdSOBA}. \emph{Our main goal is using both synthetic dataset DESOBA and rendered dataset RdSOBA to train a model, which can generate foreground shadows for real composite images.} 

Early shadow generation methods \cite{aicnet,arshadowgan} usually produce terrible results. SGRNet \cite{sgrnet} can achieve good metrics on the DESOBA test set, but the generated images have unsatisfactory visual effects and cannot generalize well to real composite images. Specifically, the generated shadows of SGRNet are prone to have notable artifacts like undesired holes and isolated fragments. We also find that SGRNet could not significantly improve the visual effects on both DESOBA and real composite images with the help of our RdSOBA dataset.

\begin{figure*}[t]
\centering
\includegraphics[width=0.9\linewidth]{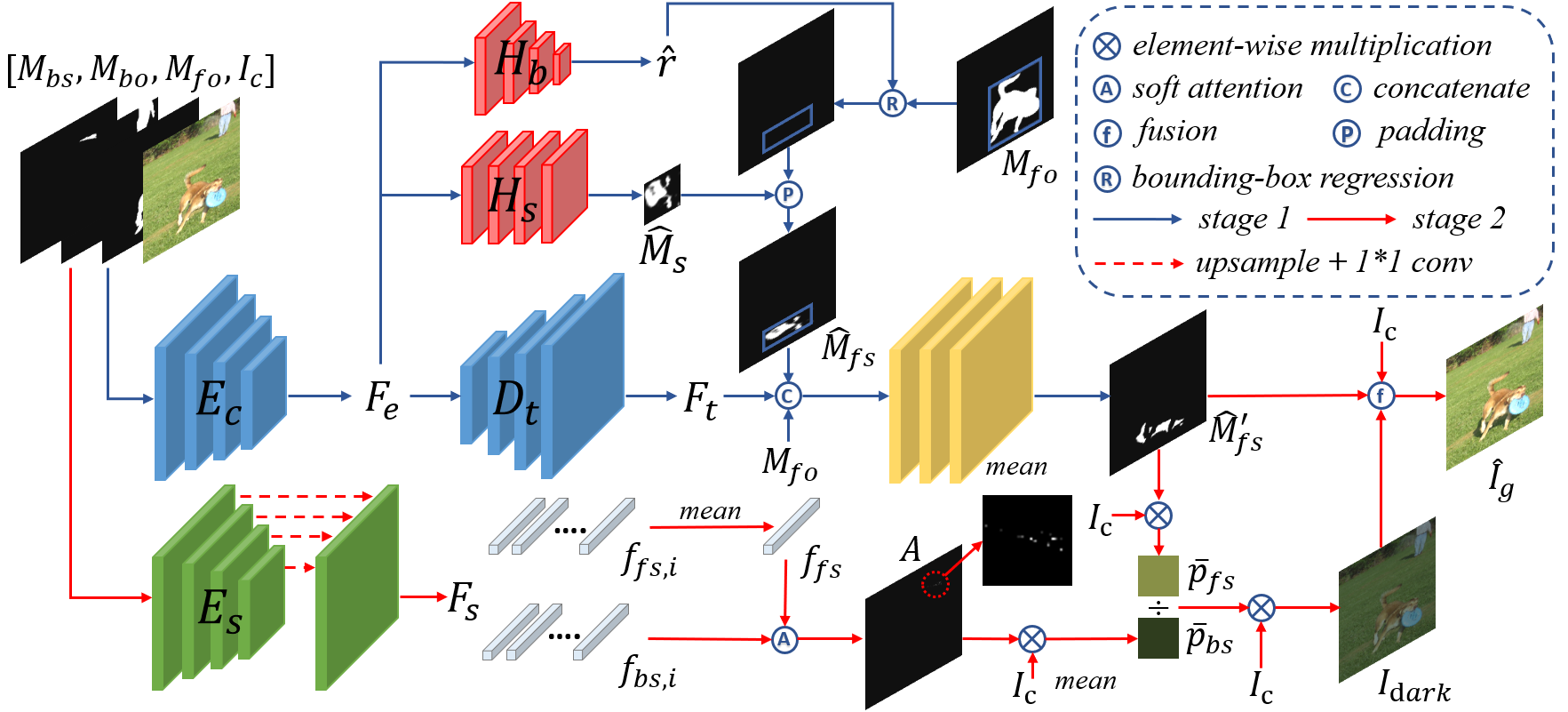}
\caption{The architecture of our proposed DMASNet. In the first stage, we employ $E_c$ to extract $F_e$, based on which the box head $H_b$ and the shape head $H_s$ jointly predict the rough mask $\hat{M}_{fs}$. By using the decoder feature from $D_t$, we refine $\hat{M}_{fs}$ to get $\hat{M}_{fs}^{'}$. In the second stage, we employ $E_s$ to extract $F_s$, based on which we calculate the attention map $A$ within background shadow region to get the target mean value for foreground shadow pixels. To match the target mean value, we scale $I_c$ to get $I_{dark}$. Finally we use $\hat{M}_{fs}^{'}$ to combine $I_{dark}$ with $I_c$ to get the final result $\hat{I}_{g}$.}
\label{fig:network}
\end{figure*}

In this work, as illustrated in Figure \ref{fig:network}, we design a two-stage network called DMASNet with \textbf{d}ecomposed \textbf{m}ask prediction and \textbf{a}ttentive \textbf{s}hadow filling. 
In the mask prediction stage, we decompose mask prediction into two sub-tasks: box prediction and shape prediction. Through this decomposition, the first (\emph{resp.}, second) sub-task focuses on predicting the scale/location (\emph{resp.}, shape) of foreground shadow. We first employ a CNN backbone to extract feature from the composite image and a set of masks, then use \textit{box head} to predict the bounding box of the foreground shadow and \textit{shape head} to predict the mask shape within the bounding box. After that, the predicted mask shape is placed in the predicted bounding box to produce a rough mask. Then, we concatenate the rough mask with the up-sampled feature map to produce a refined mask. With this structure, the predicted bounding box constrains the range of shadow to suppress isolated fragments and the shape head can generate more solid shadow by concentrating on learning the shapes of real shadows.

In the shadow filling stage, with the observation that the values of foreground shadow pixels are usually close to partial background shadow pixels, we tend to borrow relevant information from background shadow pixels. Based on multi-scale features extracted by a CNN encoder, we calculate the similarities between the average feature within foreground shadow region and pixel-wise background shadow features, resulting in an attention map within background shadow region. Then, we calculate the weighted average of background shadow pixels as the target mean value of foreground shadow pixels. Finally, we scale the pixel values within foreground shadow region to match the target mean value.
This strategy makes the generated foreground shadow look compatible with background shadows.

By treating synthetic dataset DESOBA, rendered dataset RdSOBA, and real composite images as three different domains, we also observe that our DMASNet has excellent transferability across different domains. 
With the assistance of our RdSOBA dataset, our DMASNet can achieve better visual effects than previous methods on DESOBA test set and real composite images. To sum up, our main contributions are as follows,
\begin{itemize}
\item We contribute a large-scale rendered dataset RdSOBA for the shadow generation task. 
\item We propose a novel network DMASNet with decomposed mask prediction and attentive shadow filling. 
\item Extensive experiments demonstrate that our method achieves better visual effects and generalizes well to real composite images.
\end{itemize}

\section{Related Work}
\subsection{Image Composition}
Image composition refers to cutting out a foreground object from one image and pasting it on another background image to get a composite image. The issues that make the obtained composite images unrealistic can be summarized as appearance, geometry, and semantic inconsistencies \cite{niu2021making}. Lots of works are devoted to solving one or more types of inconsistencies. For instance, object placement methods \cite{place1,place2,place3} solved the geometric inconsistency and semantic inconsistency by adjusting the scale, location, and shape of the foreground object. To address the appearance inconsistency, image harmonization methods \cite{harm1,harm2,harm3} adjusted the illumination statistics of foreground, while image blending methods \cite{blend1,blend2,blend3} attempted to blend the foreground and background seamlessly. In this work, we target at the missing foreground shadow, which also belongs to appearance inconsistency.

\subsection{Shadow Generation}
Although there are lots of works on shadow detection \cite{shadowdetect3,shadowdetect2,shadowdetect} and removal \cite{shadowremoval1,shadowremoval2,shadowremoval3,shadowremoval4}, there are few works on shadow generation. Existing works on shadow generation can be divided into two groups according to their adopted technical routes: rendering based methods and image translation methods.

\noindent\textbf{Rendering based methods:} Rendering based methods use rendering techniques to generate shadow for the foreground object, during which the 3D information of foreground object and background scene are required. With necessary 3D information, some methods \cite{render1,render2} obtained the illumination information by interacting with users, while other methods \cite{render3,render4} resorted to traditional algorithms or deep learning networks to predict additional information for a single image. Obviously, the missing information beyond a 2D image is often unavailable and hard to estimate in the real world, so the rendering based methods are difficult to use in practice.

\noindent\textbf{Image translation methods:} Deep image-to-image translation methods have become prevalent in recent years. The translation network translates a composite image without foreground shadow to a target image with foreground shadow, which can be categorized into unsupervised methods and supervised methods. Unsupervised methods like \cite{aicnet} mainly use adversarial learning to generate realistic shadows which are indistinguishable from real shadows. Supervised methods have ground-truth images with foreground shadows as supervision. \citet{shadowgan} exploited a local discriminator to finetune the details and a global discriminator to finetune the global view. \citet{arshadowgan} proposed to locate object-shadow pairs in the background to help generate the foreground shadow. \citet{SSN} predicts the ambient occlusion maps first, then uses a encoder-decoder with the help of the light map manually provided by users to generate soft shadows for the foreground object in simple backgrounds. \citet{sgrnet} designed cross-attention between foreground encoder and background encoder to predict the mask of foreground shadow, followed by filling the shadow region using predicted shadow parameters. 

In this work, we propose novel decompose mask prediction and attentive shadow filling to achieve better transferability and better visual effects.

\section{Our RdSOBA Dataset} \label{sec:RdSOBA}
In this section, we will introduce the details of constructing our RdSOBA dataset. 
\subsection{Constructing 3D Scenes}
We use Unity-3D to construct 3D scenes and render images. There are abundant freely available 3D models and 3D scenes online. We collect 800 standard 3D objects from different categories (\emph{e.g.}, people, vehicles, plants, animals) from CG websites, and 30 representative 3D scenes (\emph{e.g.}, schools, streets, grass) from Unity Asset Store and CG websites. The collected 3D objects and 3D scenes cover a wide range of 3D geometry and object/scene categories, which lays the foundation for generating diverse rendered images.

For each 3D scene, we select 20 open areas to place 3D objects, then choose 10 suitable camera settings (\emph{e.g.}, camera position and viewpoint) for each area. 
After fixing the camera, we place a group of 3D objects (randomly select 1 to 5 objects from collected 800 objects)  in the field of view of camera. For each camera setting, we place 10 groups of 3D objects.
With the camera and objects ready, we choose 5 different illumination conditions (\emph{e.g.}, illumination direction, intensity) and render a set of 2D images. Therefore, we can render $30\times 20\times 10\times 10\times 5=300,000$ sets of images. 

\subsection{Rendering 2D Images}
Once we set up an open area, camera settings, 3D object group, and lighting in a 3D scene, we generate a series of images. First, we render an image $I_{empty}$ before placing objects. After placing a group of $K$ 3D objects, we switch off the visibility of all objects to the camera and switch on the visibility of each object one by one. 
When switching on the visibility of the $k$-th object, we can choose whether to render shadow for this object in Unity-3D. Without the shadow, we render an image $I_{o,k}$ with the $k$-th object. With the shadow, we render an image $I_{os,k}$ with the $k$-th object and its shadow. Based on $I_{empty}$ and $I_{o,k}$, we can obtain the $k$-th object mask $M_{o,k}$ by calculating their difference. Similarly, based on $I_{o,k}$ and $I_{os,k}$, we can obtain the $k$-th shadow mask $M_{s,k}$. At last, we switch on the visibility of all objects and render an image $I_g$ with the shadows of all objects. 

Among  $K$ 3D objects, when we choose the $k$-th object as the foreground object, we can use $M_{o,k}$ (\emph{resp.}, $M_{s,k}$) as the foreground object (\emph{resp.}, shadow) mask $M_{fo}$ (\emph{resp.}, $M_{fs}$). We can merge 
$\{M_{o,1}, \ldots, M_{o,k-1}, M_{o,k+1}, \ldots, M_{o,K}\}$ as the background object mask $M_{bo}$. Similarly,  we can get the background shadow mask $M_{bs}$.
Then, we calculate $I_c$ by $I_c = I_{o,k} * (1 - M_{bs} - M_{bo})+ I_g * (M_{bs} + M_{bo})$, in which $*$ denotes element-wise multiplication. Up to now, we have obtained data tuples whose form is identical to DESOBA's. However, not all the tuples are of high quality because some shadow masks are incomplete or erroneous. After manually filtering out low-quality tuples, we have 280,000 tuples left. We have shown some examples in Figure \ref{fig:example} and more examples can be found in Supplementary. 

Although our dataset construction requires certain manual efforts, it is still much more efficient than manually removing the shadows when constructing DESOBA dataset. 

\section{Our Method}
Given an input composite image $I_c$, the foreground object mask $M_{fo}$, and the masks of background object-shadow pairs $\{M_{bo},M_{bs}\}$, our DMASNet aims at generating $\hat{I}_g$ with foreground shadow. We solve this problem with two independent stages: mask prediction stage and shadow filling stage. In the mask prediction stage, we decompose mask prediction into box prediction and shape prediction. The predicted mask shape is placed within the predicted bounding box to form the shadow mask $\hat{M}_{fs}$, further refined as $\hat{M}'_{fs}$. In the shadow filling stage, we predict the target mean value of foreground shadow pixels by attending to relevant background shadow pixels. After filling the predicted shadow mask to match the target mean value, we get the result $\hat{I}_g$. 

\subsection{Decomposed Mask Prediction}
In the first stage, our goal is to predict the foreground shadow mask $\hat{M}'_{fs}$. We use a CNN encoder $E_c$ to extract the bottleneck feature map $F_e$ with size $16\times 16$ from the concatenation of $\{I_c, M_{bs}, M_{bo}, M_{fo}\}$ with size $256\times 256$.

Based on $F_e$, we predict the mask shape and bounding box separately, similar to instance segmentation method Mask-RCNN~\cite{maskrcnn}. 
Unlike instance segmentation where the instance already exists in the input image, we need to imagine the shadow's shape and bounding box, which is much more challenging than instance segmentation. 

Intuitively, by observing background object-shadow pairs, we can roughly estimate the relative scale and location offset of shadow compared with the inserted object. Therefore, we predict the regression from the bounding box of foreground object to that of foreground shadow. We use a quadruplet $B=(x, y, w, h)$ to describe a bounding box,  where $(x,y)$ are the center coordinates of the bounding box and $w$ (\emph{resp.}, $h$) is the width (\emph{resp.}, height) of the bounding box. The bounding box of foreground object is denoted as $B_o=(x_o, y_o, w_o, h_o)$ and the ground-truth bounding box of foreground shadow is denoted as $B_s=(x_s, y_s, w_s, h_s)$. Then, we use another quadruplet $r=(r_x, r_y, r_w, r_h)$ to characterize the regression from $B_o$ to $B_s$:
\begin{eqnarray}\label{r}
\left \{
\begin{array}{l}
r_{x} = (x_s - x_o) / w_o, \\
r_{y} = (y_s - y_o) / h_o, \\
r_{w} = ln(w_s / w_o), \\
r_{h} = ln(h_s / h_o). \\
\end{array}
\right.
\end{eqnarray}
We use box head $H_b$ to predict quadruplet $\hat{r}$, then calculate the predicted $\hat{B}_s$ by Eqn. \ref{r}. The aspect ratio and position may change significantly when mapping the object bounding box to the shadow bounding box. To consider the coordinate correlation and enhance the robustness, we employ \textit{CIoU loss} \cite{ciou} to supervise $\hat{r}$:
\begin{eqnarray}\label{CIoU}
L_{reg}(\hat{r},r) = CIoU(\hat{B}_s,B_s).
\end{eqnarray}

For shape prediction, we do not consider position and aspect ratio handled in $H_b$, so we generate shape mask within a standardized box for each shadow. For images with size $256\times 256$, the mean size of shadow bounding boxes is close to $32\times 32$, so we use the shape head $H_s$ to predict a mask $\hat{M}_{s}$ with size $32\times 32$. We use $\xi_{B}(I)$ to denote the operation of cropping the bounding box $B$ from image $I$ and resizing it to $32\times 32$. Inversely, $\xi_{B}^{-1}(I)$ denotes the operation of placing the resized $I$ within the bounding box $B$ and padding zeros outside the bounding box. In the training phase, the ground-truth shape mask can be obtained by $M_{s} = \xi_{B_s}(M_{fs})$. We calculate $L_1$ loss between $M_{s}$ and $\hat{M}_{s}$:
\begin{eqnarray}\label{shape}
L_{shape}(\hat{M}_{s},M_{s}) = L_1(\hat{M}_{s},M_{s}).
\end{eqnarray}

After obtaining $\hat{B}_s$ and $\hat{M}_{s}$, we can get a rough shadow mask by $\hat{M}_{fs}=\xi_{\hat{B}_s}^{-1}(\hat{M}_{s})$. $\hat{M}_{fs}$ has roughly correct location, scale, and shape, but lacks fine-grained details. To compensate for detailed information, we employ a decoder $D_t$, which takes in the bottleneck feature map $F_e$ to produce the up-sampled feature map $F_t$. Then, $F_t$ is concatenated with $\{\hat{M}_{fs},M_{fo}\}$ and passed through several convolutions layers to yield the refined mask $\hat{M}'_{fs}$. Again, we use L1 loss to supervise $\hat{M}'_{fs}$:
\begin{eqnarray}\label{mask}
L_{mask}(\hat{M}'_{fs},M_{fs}) = L_1(\hat{M}'_{fs},M_{fs}).
\end{eqnarray}

\begin{figure*}[t]
\centering
\includegraphics[width=0.95\linewidth]{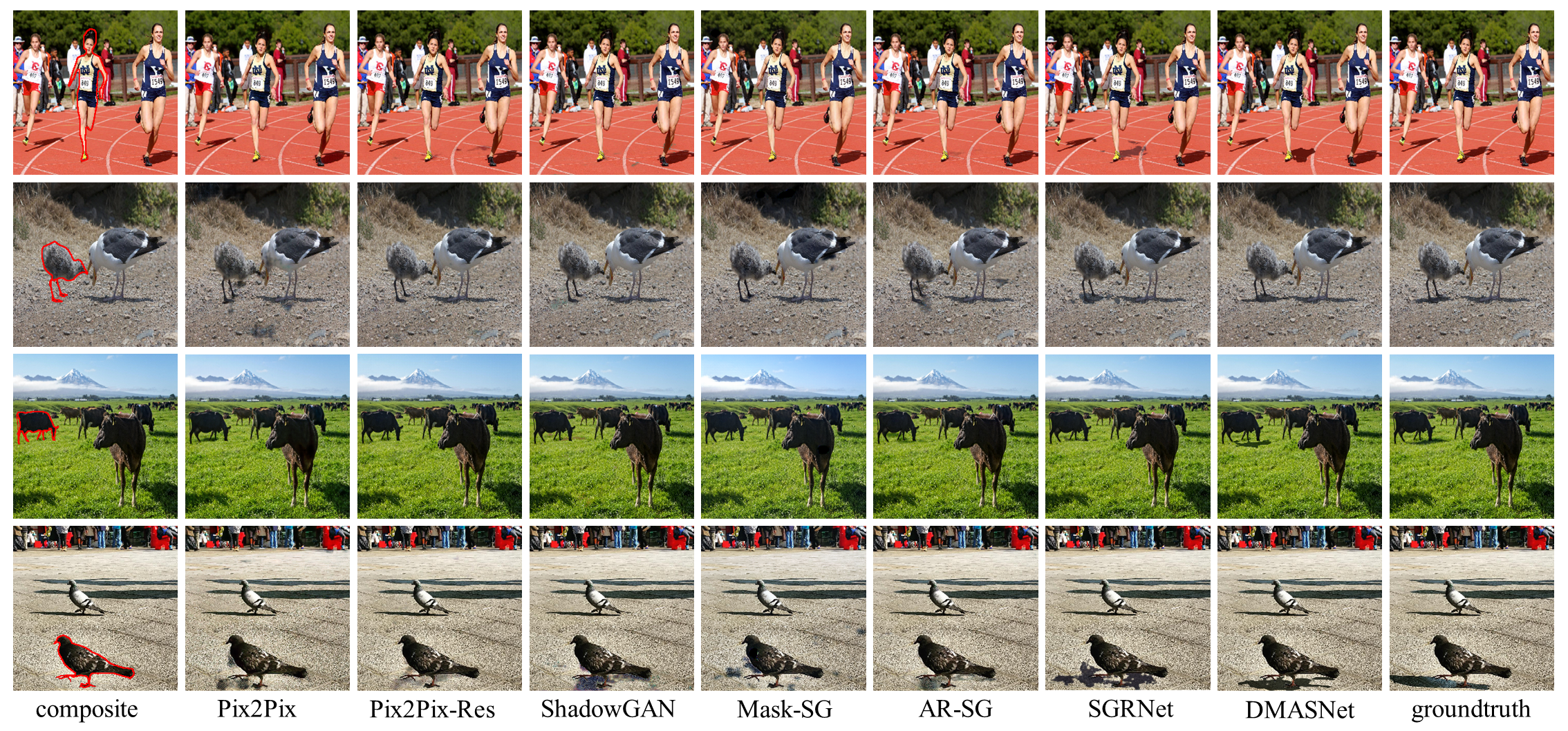}
\caption{Example results of different methods in the setting of DESOBA $\rightarrow$ DESOBA.}
\label{fig:compare1}
\end{figure*}

\subsection{Attentive Shadow Filling}
In the second stage, we aim to fill the predicted foreground shadow region. To guarantee that the filled foreground shadow looks compatible with background shadows, one naive approach is calculating the mean value of background shadow pixels as the target mean value of foreground shadow pixels. However, the values of background shadow pixels vary widely and the provided background shadow masks are not perfectly accurate. Hence, we attempt to select reference background shadow pixels by learning different weights for different background shadow pixels.

We apply another encoder $E_s$ to the concatenation of $\{I_c, M_{bs}, M_{bo}, M_{fo}\}$, and concatenate multi-scale encoder features as $F_s$. Then we average the pixel-wise feature vectors within $\hat{M}'_{fs}$ as $f_{fs}$. Besides, we denote the pixel-wise feature vector for the $i$-th background shadow pixel as $f_{bs,i}$.
After that, we project $f_{fs}$ and $f_{bs,i}$ to a common space via a fully-connected layer $\phi(\cdot)$ and calculate their similarities to produce the attention map $A=\{a_1, a_2, \ldots, a_{N_{bs}}\}$, in which $N_{bs}$ is the number of background shadow pixels and $a_i=\frac{\exp\left(\phi(f_{fs})^T\phi(f_{bs,i})\right)}{\sum_{i=1}^{N_{bs}}\exp\left(\phi(f_{fs})^T\phi(f_{bs,i})\right)}$. Finally, we calculate the weighted average of background shadow pixels as $\bar{p}_{bs} = \frac{1}{N_{bs}} \sum_{i=1}^{N_{bs}} a_i p_{bs,i}$, in which $p_{bs,i}$ is the color value of the $i$-th background shadow pixel. 

Then, we calculate the average of pixel values within the predicted foreground shadow region $\hat{M}'_{fs}$ as  $\bar{p}_{fs}$. To match the target mean value $\bar{p}_{bs}$, we can get the scale $\bar{p}_{bs}/\bar{p}_{fs}$ for the foreground shadow region. In implementation, we scale the whole composite image $I_c$ by $I_{dark} = \bar{p}_{bs}/\bar{p}_{fs} * I_c$ and obtain the target image $\hat{I}_g$ by
\begin{eqnarray}\label{mask}
\hat{I}_g = \hat{M}'_{fs} * I_{dark} + (1 - \hat{M}'_{fs}) * I_c.
\end{eqnarray}

We employ shadow-MSE loss to supervise the shadow color for the predicted target image:
\begin{eqnarray}\label{mask}
L_{rec}(\hat{I}_g,{I}_g) = MSE(\hat{I}_g*M_{fs},{I}_g*M_{fs}).
\end{eqnarray}

The whole network can be trained in an end-to-end manner, and the overall loss function can be written as
\begin{eqnarray}\label{mask}
L_{total} = L_{reg} + L_{shape} + L_{mask} + L_{rec}.
\end{eqnarray}

\section{Experiments}
\subsection{Datasets and Implementation Details}

Our RdSOBA dataset has 280,000 training tuples. DESOBA dataset has 2792 training tuples and 580 testing tuples. All images are resized to $256\times 256$. We implement our model using PyTorch and train our model on 4*RTX 3090 with batch size being 16. We use the Adam optimizer with the learning rate being 0.0001 and $\beta$ set to (0.5,0.999). We train RENOS for 50 epochs and DESOBA for 1000 epochs without using data augmentation. Further details of our network can be found in the Supplementary.

\subsection{Baselines}

Following \cite{sgrnet}, we pick Pix2Pix \cite{pix2pix}, Pix2Pix-Res, Mask-ShadowGAN~\cite{maskshadowgan}, ShadowGAN~\cite{shadowgan}, ARShadowGAN~\cite{arshadowgan}, and SGRNet~\cite{sgrnet} as baselines.  Pix2Pix is a typical image-to-image translation method. Pix2Pix-Res is a simple variant of Pix2Pix. Mask-ShadowGAN performs shadow removal and mask-guided shadow generation simultaneously, in which the shadow generation network can be adapted to our task. ShadowGAN, ARShadowGAN, and SGRNet work on the same task as ours and thus can be directly applied. 

\begin{figure*}[t]
\centering
\includegraphics[width=0.9\linewidth]{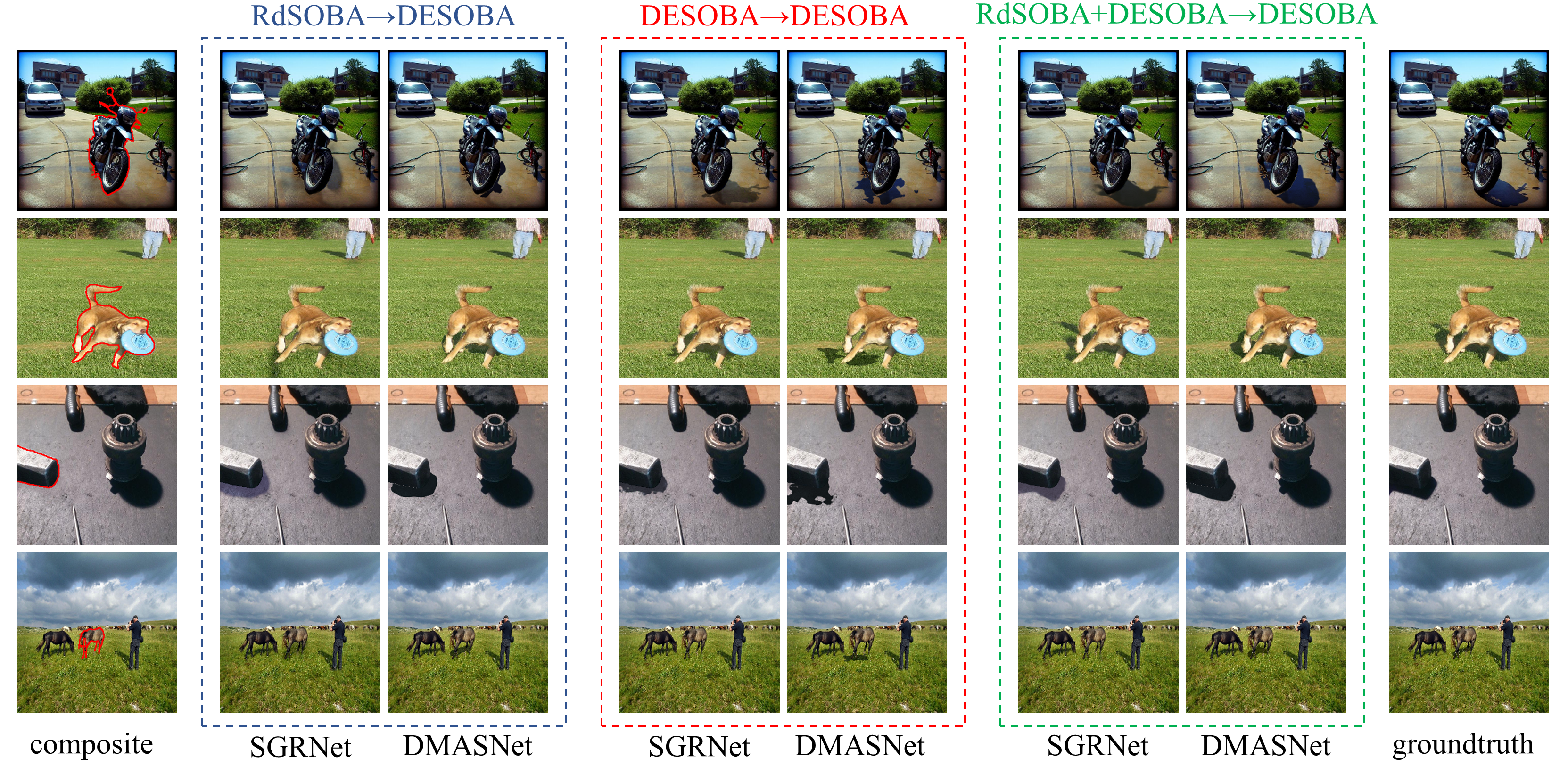}
\caption{Example results of a comprehensive comparison between our DMASNet and SGRNet.}
\label{fig:compare2}
\end{figure*}

\subsection{Evaluation Metrics}
\begin{table}[t]

\centering
\small
\begin{tabular}{c|cccc}
\hline 
Method & RMSE$\downarrow$ & S-RMSE$\downarrow$ & PSNR$\uparrow$ & S-PSNR $\uparrow$ \\
\hline 
Pix2Pix & 7.659 & 75.346 & 31.791 & 6.785 \\
Pix2Pix-Res & 5.961 & 76.046 & 35.073 & 6.678 \\
ShadowGAN & 5.985 & 78.413 & 35.391 & 6.337  \\
Mask-SG & 8.287 & 79.212 & 30.991 & 6.280 \\
AR-SG & 6.481 & 75.099 & 33.720 & 6.790 \\
SGRNet & \textbf{4.754} & \textbf{61.762} & \textbf{37.126} & \textbf{8.501}  \\
DMASNet & 5.583 & 65.847 & 36.013 & 8.029   \\
\hline 
\end{tabular}
\caption{Quantitative results of different methods in the setting of DESOBA $\rightarrow$ DESOBA. S-RMSE (\emph{resp.}, S-PSNR) means shadow-RMSE (\emph{resp.}, shadow-PSNR).}
\label{table:1}
\end{table}

\begin{table*}[tb]
\small
\centering
\begin{tabular}{c|c|cccc|cc|cc}
\hline
Training-set & Method & RMSE$\downarrow$ & S-RMSE$\downarrow$ & PSNR$\uparrow$ & S-PSNR $\uparrow$ & BER$\downarrow$ & S-BER$\downarrow$ & d\_hole & d\_frag \\
\hline 
\multirow{2}{*}{RdSOBA} &  SGRNet & 8.127 & 66.603 & 31.025 & 8.033 & 39.220 & 78.137 & 1.024 & 30.686 \\
~ & DMASNet & 7.119 & 63.693 & 33.637 & 8.347 & 28.726 & 57.046 & 0.362 & 10.619  \\
\hline
\hline 
\multirow{2}{*}{DESOBA} &  SGRNet & 4.754 & 61.762 & 37.126 & 8.501 & 26.624 & 53.407 & 15.797 & 100.440 \\
~ & DMASNet & 5.583 & 65.847 & 36.013 & 8.029 & 29.792 & 59.264 & 2.218 & 31.473  \\
\hline
\hline 
\multirow{2}{*}{RdSOBA+DESOBA} &  SGRNet & \textbf{4.676} & 59.855 & 36.898 & 8.921 & 27.233 & 54.300 & 5.276 & 56.995 \\
~ & DMASNet & 4.703 & \textbf{55.168} & \textbf{37.149} & \textbf{9.521} & \textbf{24.295} & \textbf{48.336} & 0.212 & 11.924  \\
\hline 
\end{tabular}
\caption{Quantitative results of SGRNet and our DMASNet on DESOBA test set using different training sets. S-BER means shadow-BER. We report d\_hole and d\_frag of  ground-truth foreground shadow masks (d\_hole=1.076, d\_frag=15.657) for reference. }
\label{table:2}
\end{table*}

For comprehensive comparison, we adopt three groups of evaluation metrics. 

The first group of metrics are calculated between the ground-truth and the predicted images. We adopt RMSE (\emph{resp.}, shadow-RMSE) and PSNR (\emph{resp.}, shadow-PSNR), that computed on the whole image (\emph{resp.}, ground-truth foreground shadow region). These metrics are valuable, but may not conform to human perception of visual quality. 

The second group of metrics are calculated between the ground-truth masks and the predicted masks. We adopt balanced error rate (BER) (\emph{resp.}, shadow-BER), which is computed based on the whole image (\emph{resp.}, ground-truth foreground shadow region). We observe that BER is relatively more consistent with human perception of visual quality.

In the third group, considering that undesired holes and isolated fragments seriously affect human perception without significantly affecting the abovementioned metrics
, we design two heuristic metrics to measure the quality of predicted masks. In particular, we calculate the difference between the original mask and that after filling the holes (\emph{resp.}, removing isolated fragments) using the functions in \textit{skimage.morphology}\footnote{remove\_small\_objects and remove\_small\_holes. We set connectivity=2 and area\_threshold=50}, which is referred to d\_hole (\emph{resp.}, d\_frag). To judge the quality of predicted masks, we should refer to the values of d\_hole and d\_frag of ground-truth masks. 

\begin{figure*}[t]
\centering
\includegraphics[width=0.92\linewidth]{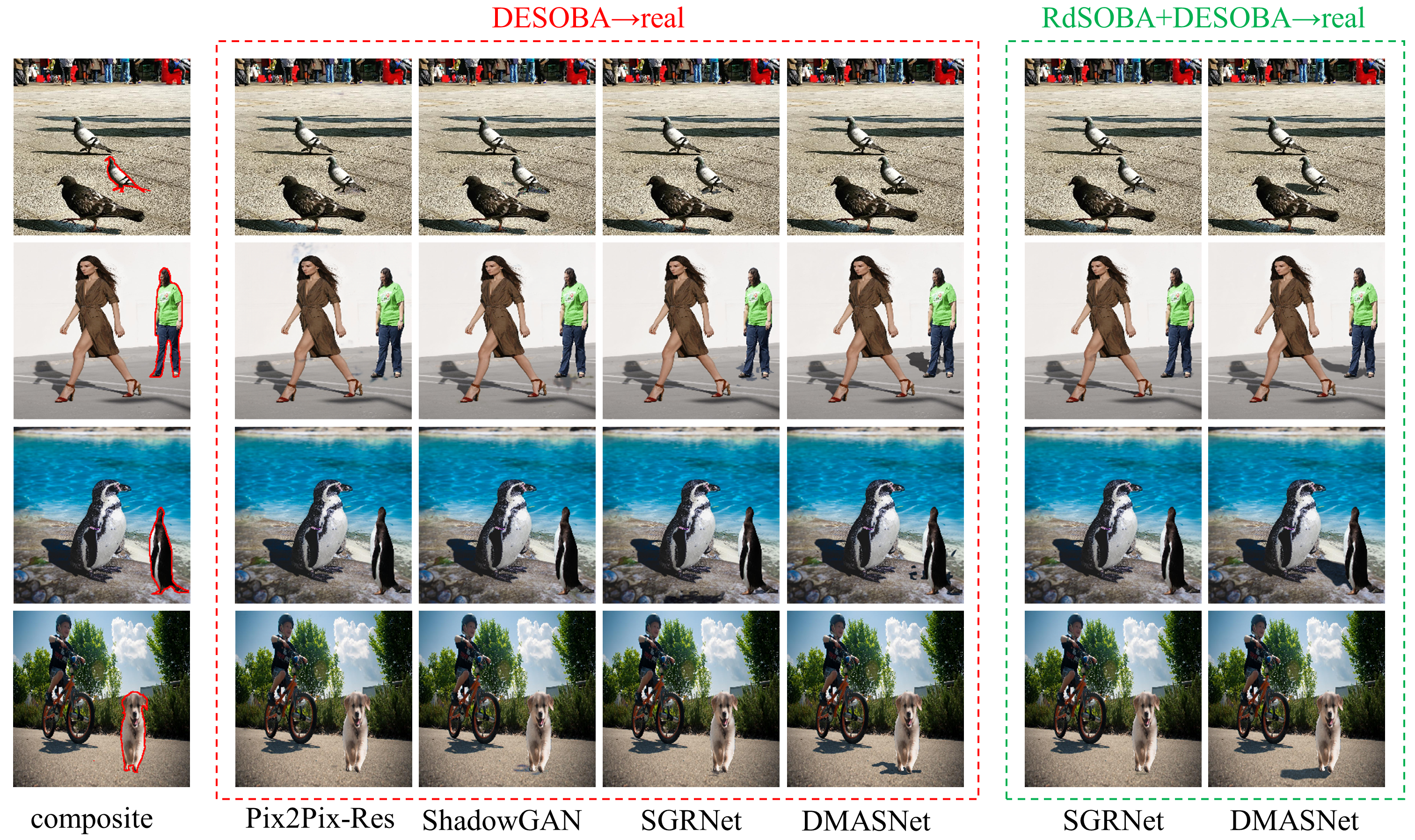}
\caption{Example results of different methods on real composite images.}
\label{fig:compare3}
\end{figure*}

\subsection{Experiments Without Using RdSOBA Dataset} \label{sec:exp_wo_RdSOBA}
In this section, we only train and evaluate different methods on DESOBA dataset. We show the qualitative results in in Figure \ref{fig:compare1} and report the quantitative results in Table \ref{table:1}. Since some baselines do not predict the foreground shadow mask, we only calculate the metrics in the first group.

As shown in Figure \ref{fig:compare1}, only our DMASNet and SGRNet can generate plausible shadows, while other methods usually generate terrible shadows. For the quantitative results in Table \ref{table:1}, although our DMASNet is worse than SGRNet on these metrics,  the quality of generated images by our DMASNet is comparable with or better than SGRNet. 
For example, our DMASNet can generate shadows with more compact shapes (\emph{e.g.}, row 4) or more proper intensity values (\emph{e.g.}, row 1). 
The mismatch between quantitative evaluation and qualitative evaluation motivates us to include more metrics for more comprehensive comparison. 

\subsection{Experiments Using RdSOBA Dataset} \label{sec:exp_with_RdSOBA}
In this section, we only compare with SGRNet because the performances of other baselines are poor as demonstrated in Section \ref{sec:exp_wo_RdSOBA}. We conduct experiments in three settings: 1) only using RdSOBA training set; 2) only using DESOBA training set; 3) using both RdSOBA training set and DESOBA training set (first training on RdSOBA, then finetune on DESOBA). For all settings, the evaluation is performed on DESOBA test set.
The full comparisons between DMASNet and SGRNet in three settings are summarized in Table \ref{table:2}. We show the advantages of our method from three aspects.

\noindent\textbf{Evaluation metrics:} With the help of our RdSOBA, our DMASNet demonstrates notable superiority over SGRNet in metrics. Nevertheless, our RMSE and PSNR remains comparable to SGRNet's. One reason is that our DMASNet is more likely to produce more shadows than SGRNet, so the MSE of those false positive pixels (non-shadow pixels that are wrongly predicted as shadow pixels) is larger. When using DESOBA or RdSOBA+DESOBA as training set, compared with SGRNet, the d\_hole and d\_frag of our DMASNet are also closer to those of ground-truth masks. 

\noindent\textbf{Cross-domain transferability:} When using RdSOBA as training set, the results of SGRNet are poor (row 1). In contrast, the results of DMASNet are even better than those obtained by using DESOBA training set (row 2 \emph{v.s.} row 4). The contrasting behaviors of SGRNet and our DMASNet show the gap between their transferability. Moreover, with RdSOBA, DMASNet can achieve much better results (row 6 \emph{v.s.} row 4), but the improvement of SGRNet is marginal (row 5 \emph{v.s.} row 3), which also demonstrates that our DMASNet has better transferability across different domains. 

\noindent\textbf{Visual effects:} As shown in Figure \ref{fig:compare2}, the results of DMASNet are substantially improved with the aid of RdSOBA, which are more realistic and closer to the ground-truth images. However, the improvement of SGRNet is not apparent.

\begin{table}[t]
\small
\centering
\begin{tabular}{c|c|ccc}
\hline 
Training-set & Method & B-T score$\uparrow$ & d\_hole & d\_frag  \\
\hline 
\multirow{4}{*}{DESOBA} & Pix2Pix-Res & -1.149 & - & - \\
~ & ShadowGAN & -1.206 & - & - \\
~ & SGRNet & -0.104 & 21.540 & 200.430 \\
~ & DMASNet & 0.648 & 2.630 & 67.300 \\
\hline
\multirow{2}{*}{\parbox{1.2cm}{RdSOBA+ \\ DESOBA}} & SGRNet & -0.192 & 5.460 & 124.050  \\
~ & DMASNet & \textbf{2.039} & 1.440 & 18.930  \\
\hline 
\end{tabular}
\caption{Quantitative results of different methods on real composite images.}
\label{table:3}
\end{table}

\subsection{Evaluation on Real Composite Images}
Our ultimate goal is generating realistic foreground shadow for real composite images, so we compare different methods on the 100 real composite images provided by \cite{sgrnet}. We compare our DMASNet with SGRNet as well as relatively strong baselines Pix2Pix-Res and ShadowGAN (see Table \ref{table:1}). For DMASNet and SGRNet, we provide two versions depending on whether using RdSOBA dataset or not. As there are no ground-truth  images, we report the metrics d\_hole, d\_frag and conduct user study.  We only report d\_hole, d\_frag for DMASNet and SGRNet, because the other two methods do not predict mask.

In our user study, we invite 20 people to observe a pair of generated images by two methods at a time and ask them to choose the one with  more realistic foreground shadow. Based on pairwise scores, we use Bradley-Terry(B-T) model~\cite{abc} to calculate the global ranking score for each method. We list the results in Table \ref{table:3} and show the generated images by different methods in Figure \ref{fig:compare3}.
We can see that our DMASNet can greatly benefit from RdSOBA and consistently produce more realistic results, while SGRNet fails to produce shadow in many cases.

\section{Conclusion}
In this work, we have contributed a large-scale rendered shadow generation dataset RdSOBA. We have also proposed a novel two-stage shadow generation network DMASNet, which decomposes mask prediction and performs attentive shadow filling. Extensive experiments have proved that our RdSOBA dataset is helpful. Our DMASNet has shown remarkable cross-domain transferability and achieved the best visual effects for real composite images.

\section*{Acknowledgements}
The work was supported by the National Natural Science Foundation of China (Grant No. 62076162), the Shanghai Municipal Science and Technology Major/Key Project, China (Grant No. 2021SHZDZX0102, Grant No. 20511100300). 

\bibliography{main, supp}

\end{document}


\maketitle

In this Supplementary, we will first introduce more implementation details of our DMASNet in Section \ref{network_detail}, then analyse cross-domain transferability in Section \ref{transferability} and conduct ablation studies in Section \ref{ablation} to demonstrate the effectiveness of our network design. Then, we will show more example tuples of our RdSOBA dataset in Section \ref{more_example}. After that, to show the advantages of our network from more aspects, we will compare the efficiency with SGRNet in Section \ref{efficiency}, and display more results on real composite images with one or two foreground objects generated by different methods in Section \ref{twofg}. Finally, we will discuss the limitations of our DMASNet in Section \ref{limit}.

\section{Details of Network Architecture} \label{network_detail}
\noindent\textbf{Encoder $E_c$:} The structure of $E_c$ is based on the ResNet34~\cite{resnet_supp}. We use instance normalization(IN) after convolution and ReLU activation function after normalization. Before entering the residual module, we use two $3\times 3$ convolution layers to replace the $7\times 7$ convolution layer in the raw ResNet34. The output feature map from $F_e$ has size $256\times 16\times 16$.

\noindent\textbf{Box head $H_b$:} We use three $3\times 3$ convolution layers (with IN and ReLU) to get a $512\times 8\times 8$ feature map from $F_e$, and then use adaptive average pooling to get a $512$-dimensional vector. Finally we use a fully connected layer to produce a $4$-dimensional vector $\hat{r}$.

\noindent\textbf{Shape head $H_s$:} We first upsample $F_e$ to $256\times 32\times 32$, and then use four $3\times 3$ convolutional layers (with IN and ReLU). Finally, we use a $3\times 3$ convolution layer with Tanh activation function to produce the $32\times 32$ output mask $\hat{M}_s$.

\begin{figure}[t]
\centering
\includegraphics[width=1.0\linewidth]{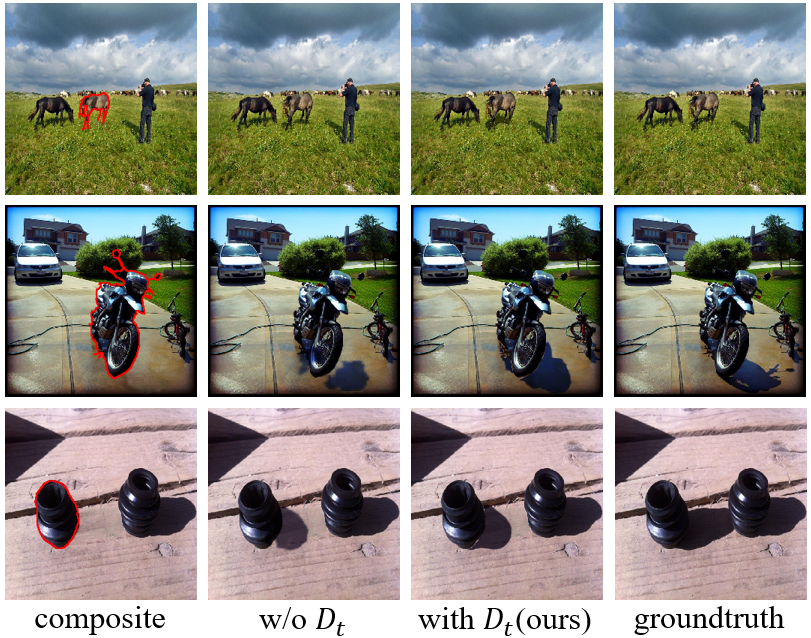}
\caption{Examples results to show the importance of $D_t$. From left to right, we show the composite image, the result of our method without using $D_t$ or using $D_t$, and the ground-truth target image. }
\label{fig:ablation1}
\end{figure}

\noindent\textbf{Refinement module $D_t$:} First, we define ``Up+Conv" block which consists of an upsampling layer and two $3\times 3$ convolution layers (with IN and ReLU). We have 4 Up+Conv in $D_t$.  After the last Up+Conv's upsample layer, we get $F_t$, which is concatenated with $\{\hat{M}_{fs}, M_{fo}\}$ and passed through the remaining layers. Finally we use a $3\times 3$ convolution layer (with IN and ReLU) and a $3\times 3$ convolution layer with Tanh to produce the final $256\times 256$ foreground mask $\hat{M}'_{fs}$.

\noindent\textbf{Encoder $E_s$:} $E_s$ is also based on the ResNet34. For the features of different scales after each residual layer, we upsample them to $256\times 256$ and apply $1\times 1$ convolution layer to align their channel numbers to be $32$. Then we concatenate these feature maps and perform $3\times 3$ convolution, and finally get the feature map $F_s$ of size $32\times 256\times 256$. 

\begin{table}[t]
\small
\begin{center}
\begin{tabular}{c|cc}
\hline 
Method & box(IoU)$\uparrow$ & shape(L1)$\downarrow$  \\
\hline 
SGRNet & 0.2317 & 0.4039  \\
ours & \textbf{0.3612} & \textbf{0.3427}  \\
\hline 
\end{tabular}
\end{center}
\caption{Quantitative results of SGRNet and our DMASNet for cross-domain transferability in the setting of RdSOBA+DESOBA $\rightarrow$ DESOBA.}
\label{table:cross-domain}
\end{table}

\begin{table}[t]
\small
\begin{center}
\resizebox{\linewidth}{!}{
\begin{tabular}{c|cccc}
\hline 
Method & BER$\downarrow$ & S-BER$\downarrow$ & d\_hole & d\_frag  \\
\hline 
w/o $D_t$ & 29.527 & 59.152 & 3.374	& 9.273  \\
with $D_t$(ours) & \textbf{24.295} & \textbf{48.336} & 0.212 & 11.924  \\
\hline 
\end{tabular}}
\end{center}
\caption{Quantitative results of whether using $D_t$ in our DMASNet in the setting of RdSOBA+DESOBA $\rightarrow$ DESOBA. }
\label{table:ablation1}
\end{table}

\begin{table}[t]
\small
\begin{center}
\resizebox{\linewidth}{!}{
\begin{tabular}{c|cccc}
\hline 
Method & RMSE$\downarrow$ & S-RMSE$\downarrow$ & PSNR$\uparrow$ & S-PSNR $\uparrow$  \\
\hline 
average & 4.871 & 56.259 & 36.973 & 9.421  \\
linear(w/o bias) & 4.932 & 57.135 & 36.812 & 9.148  \\
linear(with bias) & 4.898 & 56.623 & 36.901 & 9.382  \\
ours & \textbf{4.703} & \textbf{55.168} & \textbf{37.149} & \textbf{9.521}  \\
\hline 
\end{tabular}}
\end{center}
\caption{Quantitative results of using different color mapping strategies in the shadow filling stage. S-RMSE (\emph{resp.}, S-PSNR) means shadow-RMSE (\emph{resp.}, shadow-PSNR).}
\label{table:ablation2}
\end{table}

\begin{table*}[t]
\small
\begin{center}
\resizebox{\linewidth}{!}{
\begin{tabular}{c|cccc|cc|cc}
\hline
source of $M_{bo}$,$M_{bs}$& RMSE$\downarrow$ & S-RMSE$\downarrow$ & PSNR$\uparrow$ & S-PSNR $\uparrow$ & BER$\downarrow$ & S-BER$\downarrow$ & d\_hole & d\_frag \\
\hline 

ground-truth  & \textbf{4.703} & \textbf{55.168} & \textbf{37.149} & \textbf{9.521} & \textbf{24.295} & \textbf{48.336} & 0.212 & 11.924 \\
predict & 4.821 & 56.732 & 36.913 & 9.212 & 24.727 & 49.183 & 0.426 & 12.018  \\
\hline 
\end{tabular}}
\end{center}
\caption{Quantitative results of using ground-truth or predicted background object-shadow masks in the setting of RdSOBA+DESOBA $\rightarrow$ DESOBA.}
\label{table:ablation3}
\end{table*}

\begin{figure*}[t]
\centering
\includegraphics[width=1.0\linewidth]{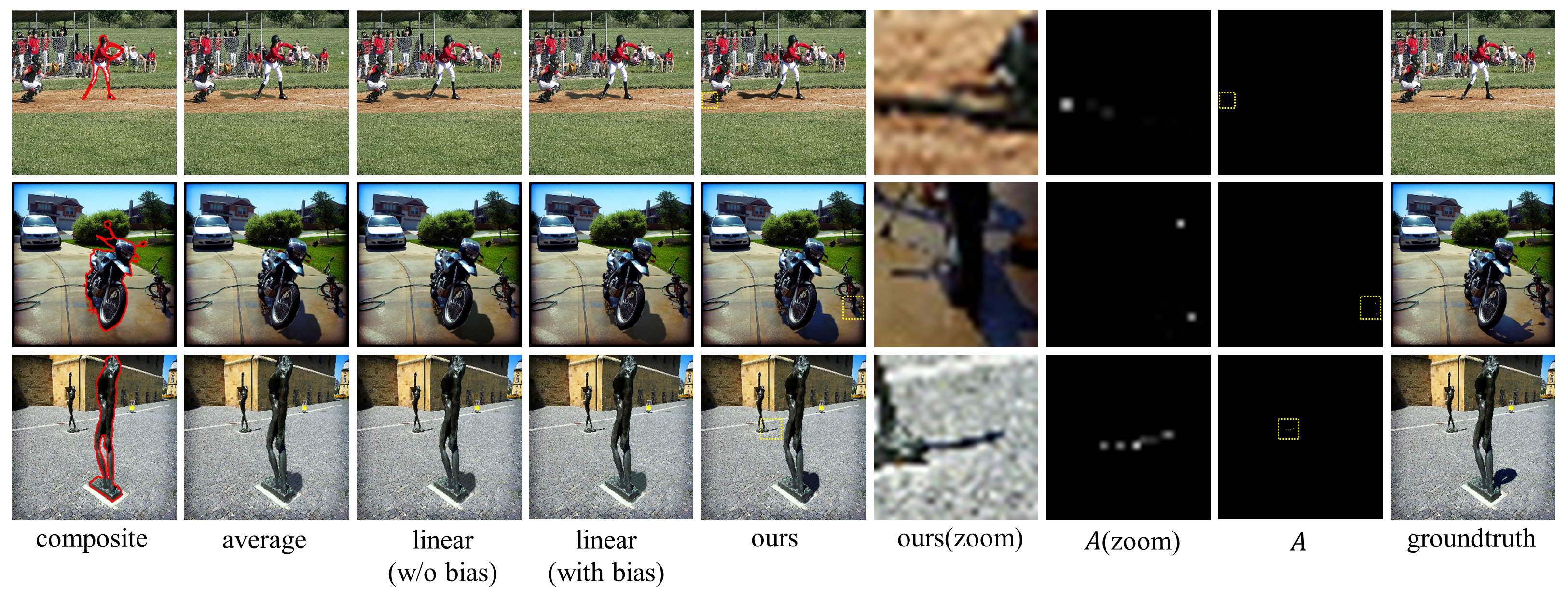}
\caption{Example results to show the advantage of our attentive shadow filling. \textit{average} means directly using the mean value of background shadow pixels as the target mean value. \textit{linear} means using an encoder to predict the linear transformation parameters in \cite{sgrnet_supp}. $A$ is the attention map learned by our model. For better observation, we zoom in the yellow dashed box in ``ours" (\emph{resp.}, ``$A$") and show it as ``ours(zoom)" (\emph{resp.}, ``$A$(zoom)").}
\label{fig:ablation2}
\end{figure*}

\section{Cross-Domain Transferability Analysis}
\label{transferability}
To provide a more intuitive comparison of the cross-domain transfer capabilities between our DMASNet and SGRNet, we evaluate their ability to predict the bounding box and shape of the foreground shadow in the setting of RdSOBA+DESOBA $\rightarrow$ DESOBA. To evaluate the quality of bounding box, we calculate the IoU between the predicted and ground-truth bounding boxes. To evaluate the quality of shadow shape, we resize the bounding box of shadow mask to $32\times 32$, and calculate $L_1$ loss between predicted shadow mask and ground-truth shadow mask. The results are summarized in Table \ref{table:cross-domain}. Our network significantly outperforms SGRNet in both bounding box and shadow shape, which validates that the ability to predict bounding box and shadow shape is transferrable from RdSOBA to DESOBA. Furthermore, these metrics are consistent with the performance of the entire shadow mask and visual effects.

\section{Ablation studies}\label{ablation}
\noindent\textbf{Mask prediction stage:} Since box head and shape head are indispensable, we mainly prove the importance of refining $\hat{M}_{fs}$ by using the upsampled feature from $D_t$. By using both RdSOBA training set and DESOBA training set, we directly use the coarse mask $\hat{M}_{fs}$ as the final mask, without refinement process. Since we only compare the first stage here, we only calculate the second and third groups of metrics. 

We list the metrics in Table \ref{table:ablation1}, which shows that BER and shadow-BER drop sharply without refinement. 
We also show some visualization results in Figure \ref{fig:ablation1}. It can be seen that the coarse masks have roughly correct locations and shapes (column 2), but the refined masks have more accurate shapes, which are closer to the ground-truth.

\begin{figure*}[!htp]
\centering
\includegraphics[width=0.9\linewidth]{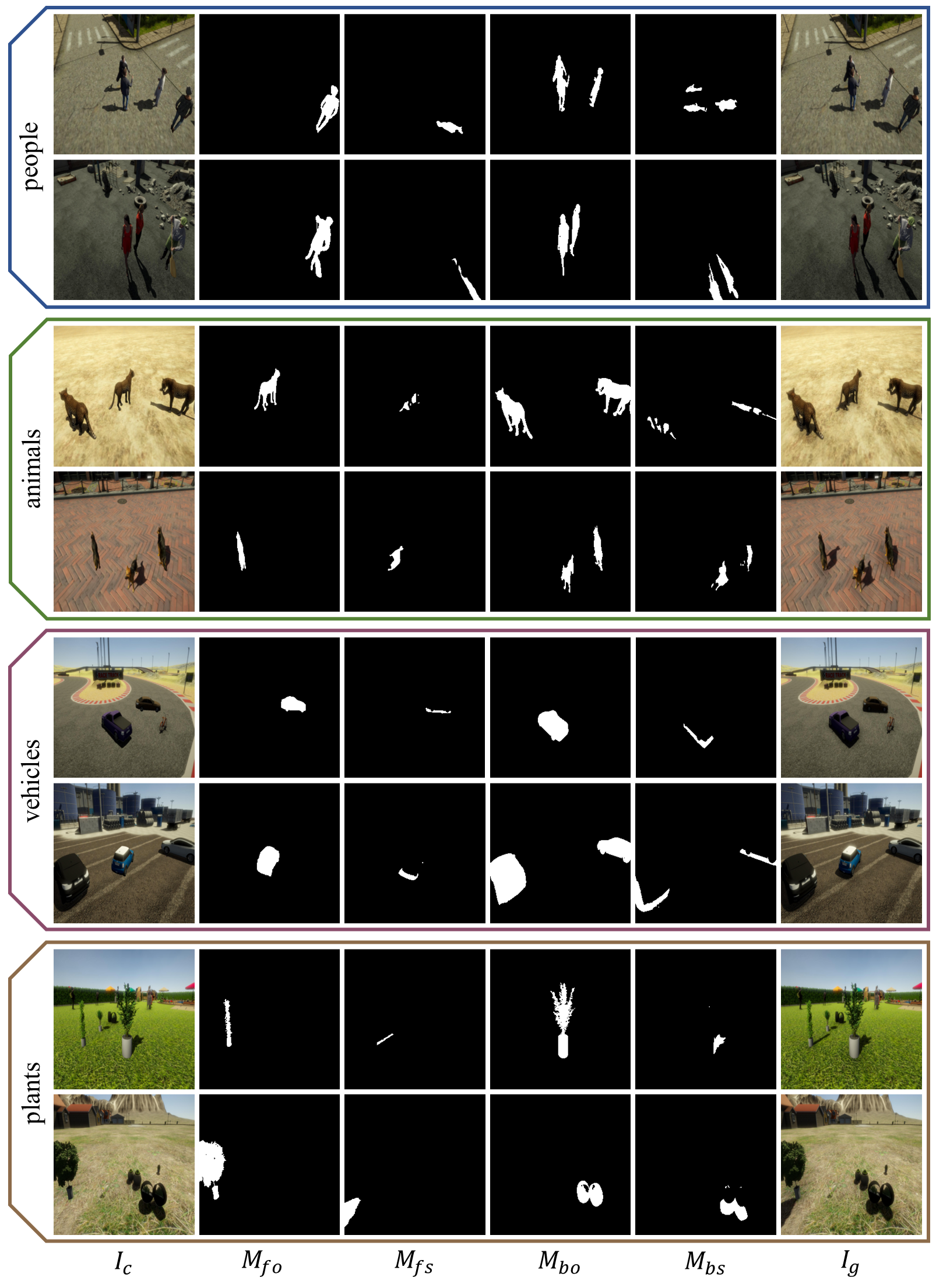}
\caption{Some examples of our RdSOBA dataset.}
\label{fig:moreRdSOBA}
\end{figure*}

\begin{figure*}[!htp]
\centering
\includegraphics[width=0.96\linewidth]{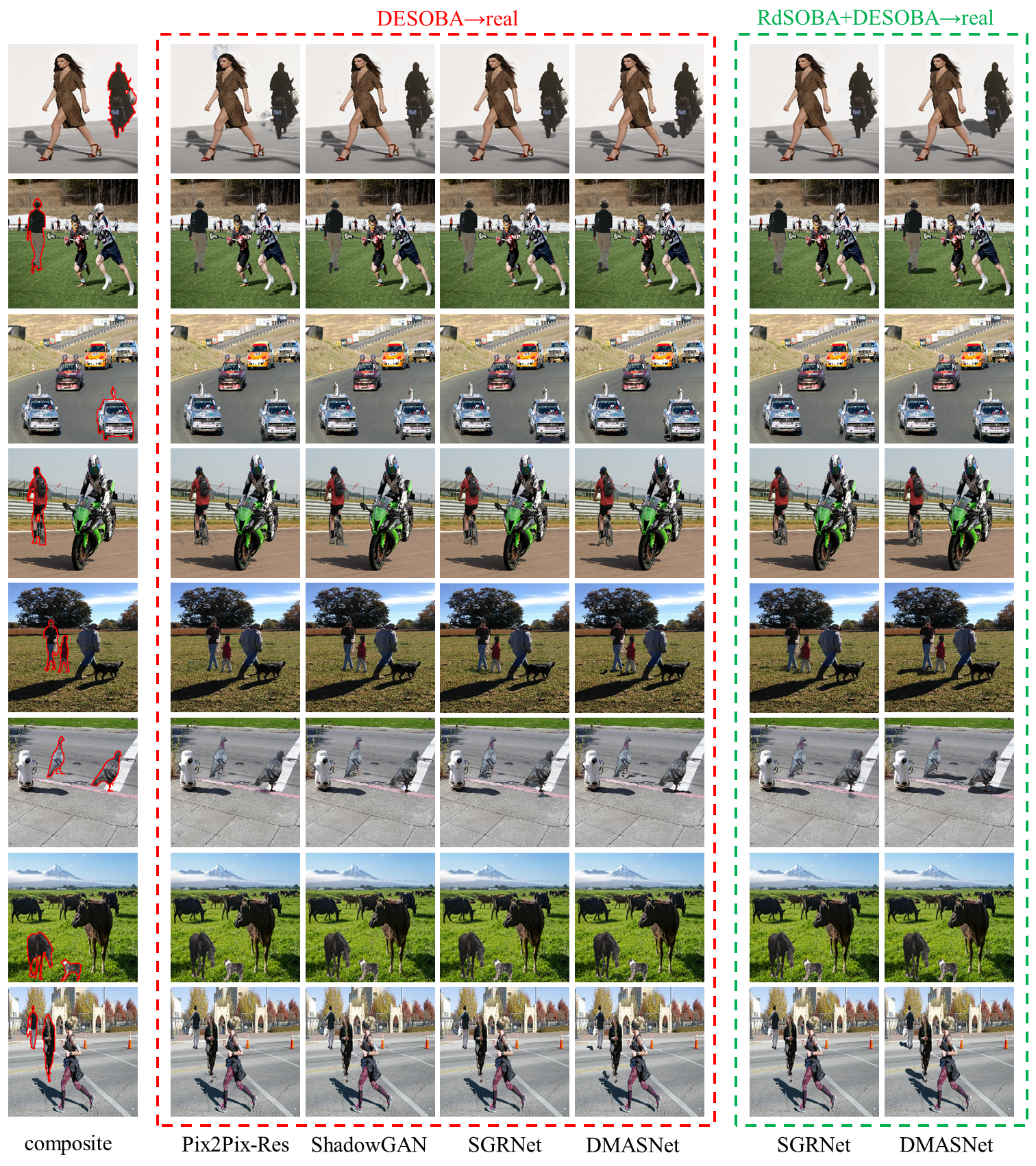}
\caption{More results of different methods on real composite images. The composite images from row 1 to row 4 have one foreground object; besides, the composite images from row 5 to row 8 have two foreground objects.}
\label{fig:twofg}
\end{figure*}

\noindent\textbf{Shadow filling stage:} 
In this subsection, we fix the model parameters in the shadow prediction stage trained using both RdSOBA and DESOBA training set. We only compare the results for the first group of metrics.

To corroborate the superiority of our attention mechanism, we compare our method with simply taking the mean value of background shadow pixels as the target foreground mean value, which is referred to as \textit{average}.
In addition, to validate the effectiveness of borrowing information from background shadow pixels, we directly predict the linear transformation parameters like \cite{sgrnet_supp}, and further divided into two models according to whether using a bias, which are referred to \textit{linear(w/o bias)} and \textit{linear(with bias)}. Specifically, we feed the concatenation of $\{M_{bs}, M_{bo}, M_{fo}, I_c\}$ to a ResNet34 encoder to predict 
these parameters. For both these methods, we scale(and maybe an additional bias) to the pixels values within the predicted foreground shadow region to match the target mean value, which is the same as our method.
We list the metrics in Table \ref{table:ablation2}.

Some example results are shown in Figure \ref{fig:ablation2}. We also visualize the attention map $A$ learned by our model, which indicates the background shadow pixels we attend to. 
We can observe that when the boundary of background shadow has a transition effect (row 1), \textit{average} is prone to produce light shadow, while our method is able to produce a reasonably dark show. Additionally, when different background shadows have notable color discrepancy (the car shadow and the bicycle shadow in row 2), \textit{average} can not obtain suitable shadow color, while our method can attend to relevant background shadow (see $A$(zoom)) to obtain suitable shadow color. Another observation is that \textit{linear(w/o bias)} and \textit{linear(with bias)} often predict incorrect target value and generate poor results.

\noindent\textbf{Without ground-truth background object-shadow masks:} In practice, the ground-truth background object-shadow masks,  \{$M_{bo}$,$M_{bs}$\} are not always available.
For generality, we adopt the existing object-shadow pair detection method \cite{shadowdetect_supp} to predict background object-shadow masks, which are used to replace the ground-truth background object-shadow masks in the input. 

According to the comparison results in Table \ref{table:ablation3}, when using the predicted background object-shadow masks, the performance of our network does not suffer a lot, proving the stability of our DMASNet in the absence of ground-truth background object-shadow masks.

\section{More Examples of Our RdSOBA Dataset} \label{more_example}
As mentioned in Section 3.1 in the main text, we collected 30 representative 3D scenes (\emph{e.g.}, schools, streets, grass) and 800 standard 3D objects from different categories (\emph{e.g.}, people, vehicles, plants, animals) from CG websites. We provide more examples from our RdSOBA dataset in Figure \ref{fig:moreRdSOBA}. For each super category of foreground objects (people, animals, vehicles, plants), we show two tuples in the form of $\{I_c,M_{fo},M_{fs},M_{bo},M_{bs},I_g\}$ in different 3D scenes.

\section{Efficiency comparison}
\label{efficiency}
Since we have shown that the performances of other baselines are poor, we only compare the efficiency between our DMASNet and SGRNet. Following \cite{sgrnet_supp}, we use \textit{torchstat} to calculate the number of parameters of the networks. In addition, we compare the inference time of the two methods. From Table \ref{table:4}, we can see that our DMASNet has fewer model parameters and a faster inference speed.

\begin{table}[t]
\small
\begin{center}
\begin{tabular}{c|cc}
\hline 
Method & \#Parameters(M)$\downarrow$ & Time(ms)$\downarrow$ \\
\hline 
SGRNet & 20.32 & 43.79 \\
DMASNet & \textbf{10.97} & \textbf{34.85} \\
\hline 
\end{tabular}
\end{center}
\caption{Efficiency comparison between SGRNet and our DMASNet.}
\label{table:4}
\end{table}

\section{More Results on Real Composite Images} \label{twofg}
We target at image composition in real-world applications, and the main goal of this work is using both rendered dataset RdSOBA and synthetic dataset DESOBA to improve the performance of shadow generation for real composite images. Therefore, we show more visualization results on real composite images in Figure \ref{fig:twofg}. It can be seen that our DMASNet can produce much better results than the other methods in the setting of RdSOBA+DESOBA$\rightarrow$real. For all the examples, with the help of RdSOBA, our DMASNet can generate relatively complete and reasonable shadows, whereas other methods are struggling to produce shadows. 

Note that in Figure \ref{fig:twofg},  we also show the results for the composite images with two foreground objects (row 5-8).
Our DMASNet can only generate shadow for one foreground object each time. For the composite images with multiple foreground objects, we can simply generate multiple foreground shadows by going through the network multiple times. As demonstrated in Figure \ref{fig:twofg}, for real composite images with multiple foreground objects, our method retains all the advantages of generating shadows for the composite images with one foreground object, whereas other methods are still struggling to produce shadows. Again, with the help of RdSOBA, our DMASNet achieves significantly better results than the other methods.

\begin{figure}[t]
\centering
\includegraphics[width=1.0\linewidth]{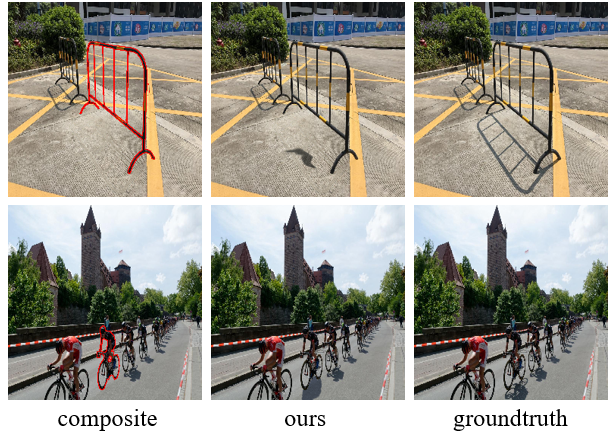}
\caption{Example failure cases of our DMASNet in the setting of RdSOBA+DESOBA $\rightarrow$ DESOBA.}
\label{fig:limitation}
\end{figure}

\section{Limitations} \label{limit}
We observe that our network can generate reasonable shadows for commonly seen objects, but it is still very challenging to generate shadows for the objects with lots of shape details. Some examples are shown in Figure \ref{fig:limitation}. The shadow shapes of the railing (row 1) and the bike (row 2) are very complicated with many details, in which cases our DMASNet fails to generate plausible results.

\bibliography{supp}